\documentclass[10pt,twocolumn,letterpaper]{article}

\usepackage[pagenumbers]{cvpr} 

\usepackage{graphicx}
\usepackage{amsmath}
\usepackage{amssymb}
\usepackage{booktabs}
\usepackage{amsmath}
\usepackage{pifont}
\usepackage{makecell}
\usepackage{multirow}
\usepackage{caption}
\usepackage{subcaption}
\usepackage{tabu}
\usepackage{makecell}

\usepackage[accsupp]{axessibility}

\usepackage[symbol]{footmisc}

\usepackage[pagebackref,breaklinks,colorlinks]{hyperref}

\usepackage[capitalize]{cleveref}
\crefname{section}{Sec.}{Secs.}
\Crefname{section}{Section}{Sections}
\Crefname{table}{Table}{Tables}
\crefname{table}{Tab.}{Tabs.}

\def\cvprPaperID{6971} 
\def\confName{CVPR}
\def\confYear{2022}

\begin{document}

\title{Distribution-Aware Single-Stage Models for Multi-Person 3D Pose Estimation}

\author{\stepcounter{footnote}\normalsize{Zitian~Wang$^{1,3}$}
\qquad \normalsize{Xuecheng~Nie$^{3}$} \qquad \normalsize{Xiaochao~Qu$^3$} \qquad \normalsize{Yunpeng~Chen$^3$} \qquad \normalsize{Si~Liu$^{1,2,}$\thanks{Corresponding author.}} \\
    \small{$^{1}$Institute of Artificial Intelligence, Beihang University, China} \\
	\small{$^{2}$Hangzhou Innovation Institute, Beihang University, China} \\
	\small{$^{3}$MT Lab, Meitu Inc., China} \\
	{\small\tt wangzt.kghl@gmail.com} \ \ {\small \tt $\{$nxc, qxc, cyp5$\}$@meitu.com}  \ \ {\small\tt liusi@buaa.edu.cn}
}

\maketitle

\begin{abstract}
In this paper, we present a novel Distribution-Aware Single-stage (DAS) model for tackling the challenging multi-person 3D pose estimation problem. Different from existing top-down and bottom-up methods, the proposed DAS model simultaneously localizes person positions and their corresponding body joints in the 3D camera space in a one-pass manner. This leads to a simplified pipeline with enhanced efficiency. In addition, DAS learns the true distribution of body joints for the regression of their positions, rather than making a simple Laplacian or Gaussian assumption as previous works. This provides valuable priors for model prediction and thus boosts the regression-based scheme to achieve competitive performance with volumetric-base ones. Moreover, DAS exploits a recursive update strategy for progressively approaching to regression target, alleviating the optimization difficulty and further lifting the regression performance. DAS is implemented with a fully Convolutional Neural Network and end-to-end learnable. Comprehensive experiments on benchmarks CMU Panoptic and MuPoTS-3D demonstrate the superior efficiency of the proposed DAS model, specifically 1.5x speedup over previous best model, and its stat-of-the-art accuracy for multi-person 3D pose estimation. 

\end{abstract}

\section{Introduction}
\label{sec:intro}

Estimating 3D poses of multiple persons from a single RGB image is a fundamental yet challenging task in computer vision, which targets at localizing 3D positions of persons and their body joints in the camera space. Recently, it has drawn much attention thanks to the broad applications in AR/VR~\cite{lin2010augmented,belghit2018vision}, Gaming~\cite{suma2011faast,shotton2011real,shotton2012efficient}, Human-Computer Interaction~\cite{erol2007vision,song2012continuous}, etc.

Prior works address this task via two-stage strategies. They either adopt the \emph{top-down} scheme~\cite{sun2018integral,simplebaseline,lcrnet,3Dposenet} that first localizes the absolute 3D person positions and then separately estimates the root-relative body joints for each person; or the \emph{bottom-up} scheme~\cite{ORPM,VoluHeatmap,SMAP} that in the first stage detects all 3D body joints and groups them into the corresponding persons in the second stage. Although achieving good accuracy, these methods suffer redundant computation and complex postprocessing, caused by the sequential management for person position and body joint localization in a two-stage manner. This leads to unsatisfied efficiency for 
deployment in real scenarios.

\begin{figure}[t!]
	\centering
	\includegraphics[width=0.5\textwidth]{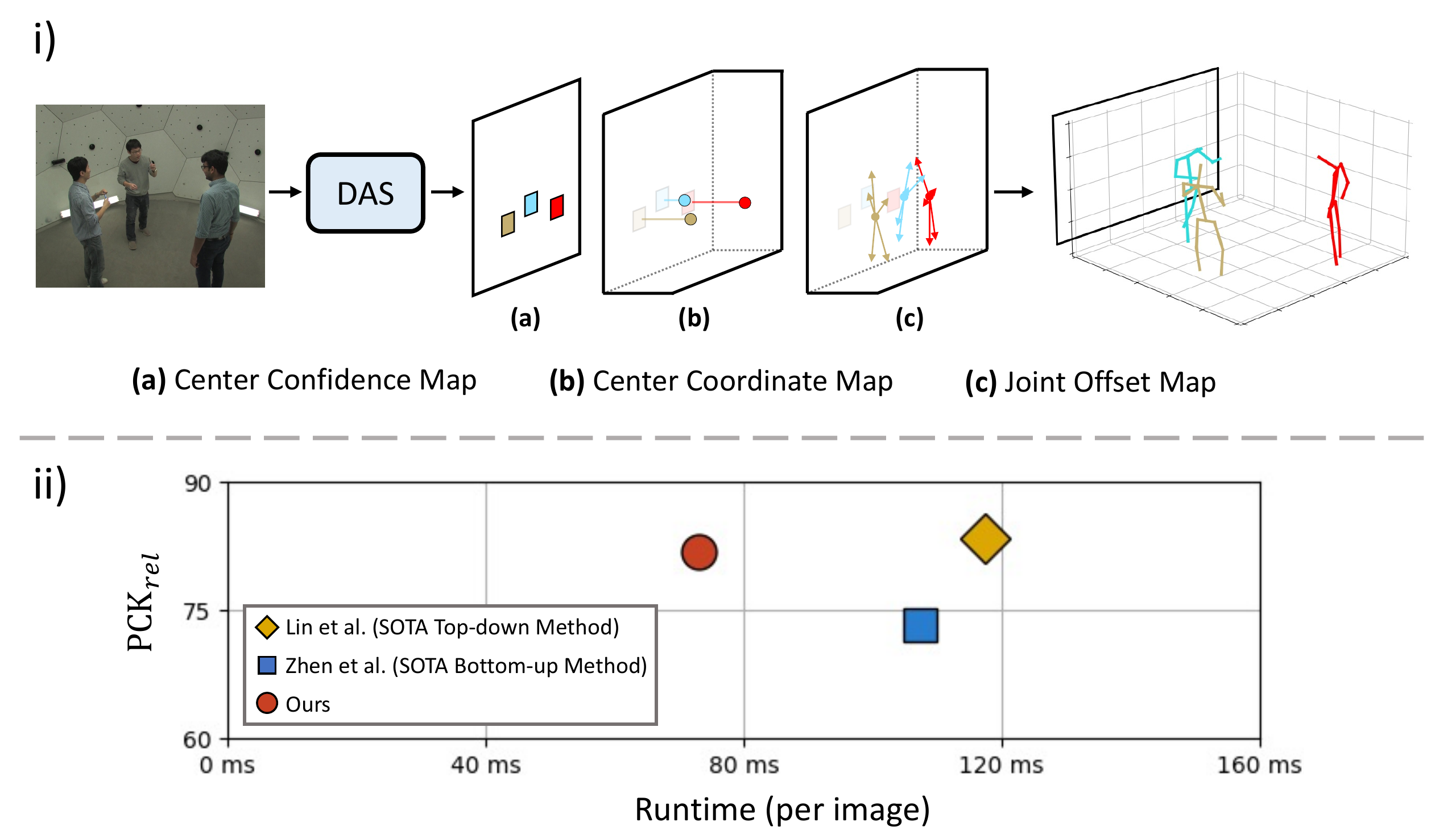}
	\vspace{-6mm}
	\caption{Overview of our Distribution-Aware Single-stage (DAS) model for multi-person 3D pose estimation. 
	i) The brief pipeline of DAS.
	ii) Comparison with state-of-the-arts on MuPoTs-3D.  DAS outperforms two-stage models in efficiency. It achieves superior performance with SOTA bottom-up method while competitive with SOTA top-down one. For PCK$_{rel}$, the higher, the better.
	}
	\label{fig:figure1}
	\vspace{-5mm}
\end{figure}

Given the above motivation, we propose to simplify the pipeline for multi-person 3D pose estimation, further pushing forward the frontier of its applications. Inspired by schemes in the 2D counterpart~\cite{centernet,spm,pointsetanchor}, we aim to design a single-stage model to simultaneously localize 3D person positions and body joints. However, the extension from 2D to 3D scenes is non-trival, due to the ill-posed setup for deriving depth information from a monocular RGB image without prior knowledge on the data distribution. 

To achieve the goal, we present a novel  Distribution-Aware Single-stage (DAS) model in this paper. The proposed DAS tackles the ill-posed problem of multi-person 3D pose estimation from two aspects: 1) DAS represents the 3D pose with a 2.5D human center together with 3D center-relative joint offsets. This adapts to direct depth prediction from the image domain as well as enables to unify the 3D localization of person position and body joints, making the monocular-based one-pass solution feasible; 2) DAS learns the true distribution of body joints during the model optimization. This provides valuable guidance to predict the joint position, thus boosting the performance of the regression-based scheme. To alleviate the distribution estimation difficulty, DAS exploits a recursive update strategy to progressively approach to the targets.
In this way, DAS can efficiently generate accurate 3D poses of multiple persons from a single RGB image.

In particular, we build DAS with a regression-based pipeline that outputs the 3D human poses via a single forward inference from an input image. As depicted in Figure~\ref{fig:figure1} i) (a) and (b), DAS models the human center with a  center confidence map and a center coordinate map. DAS uses the center confidence map for localizing the projected human centers in 2D image coordinate space, while the center coordinate map for pixel-wisely estimating the absolute center positions in 3D camera coordinate space. DAS utilizes joint offset maps to densely encode 3D center-relative locations of body joints, as shown in Figure~\ref{fig:figure1} i) (c). DAS can produce these three kinds of maps in parallel and then easily reconstruct multiple 3D human poses with them, avoiding redundant computation and complex association. With this compact single-stage pipeline, DAS can achieve superior efficiency over prior two-stage ones.

To optimize the regression-based model, prior works always adopt the conventional L1 or L2 loss. However, researches~\cite{RLE} have proven that this kind of supervision actually makes a simple Laplacian or Gaussian assumption on the data distribution, which is far way from the true one. In contrast, DAS learns the underlying distribution of 3D body joint locations via exploiting the normalizing flows~\cite{agnelli2010clustering,rezende2015variational,NICE}. This helps to derive a suitable distribution for model output, thus providing valuable priors to guide the regression of body joint coordinates. Together with the pose regression modules, DAS optimizes the distribution learning module through the maximum likelihood estimation in the training phase, while removes it in the inference phase. In this way, DAS boosts the regression performance without additional computation cost. Besides, DAS iteratively updates joint offsets by ensemble of informative predictions around regression targets to further facilitate the localization of body joints. With this distribution-aware design, DAS can achieve outperforming accuracy with bottom-up methods while competitive with top-down ones, as shown in Figure~\ref{fig:figure1} ii).

We implement DAS with a fully Convolutional Neural Network, which is end-to-end learnable. Comprehensive experiments on benchmarks CMU Panoptic~\cite{cmupanoptic} and MuPoTS-3D~\cite{ORPM} show the superiority of our DAS model. In summary, our contribution are in three folds: (1) We present a novel single-stage model for estimating 3D poses of multiple persons from a monocular RGB image, which overcomes the drawbacks on computation cost and model complexity occurred in two-stage methods; (2) We present a novel distribution-aware way to boost the body joint regression in a recursive manner; (3) We set new state-of-the-art on multiple benchmarks with superior efficiency.

\section{Related Works}

\subsection{Two-stage 3D Human Pose Estimation}
Approaches for monocular multi-person 3D pose estimation can be divided into two categories, the top-down methods and bottom-up methods. 
The top-down methods require additional human detector to provide human position prior for downstream single person pose estimator to generate individual 3D poses~\cite{pavlakos2017coarse,sun2017compositional,sun2018integral,simplebaseline}. 
\cite{lcrnet,lcrnet++} introduce a pose proposals network to generate human bounding boxes and a series of human pose hypotheses, then the pose hypotheses are refined based on the cropped ROI features. 
To estimate the camera-centric human pose, \cite{3Dposenet} disentangles absolute human depth estimation and human-centric pose estimation by using separate models. 
\cite{HMOR} propose hierarchical ordinal relations objective along with virtual view sample strategy to constrain the instance level and joint level depth values.
 \cite{HDNet} utilize GNN to infer the human depth based on the human-centric joint features.
Although achieving high accuracy, top-down methods suffers high computation cost as person number increases.

Another group of approaches investigate multi-person 3D pose estimation in a bottom-up manner.
Similar to the case in 2D pose estimation, the bottom-up methods are commonly composed of joint localization and joint association.
\cite{ORPM} combines the joint location maps and the occlusion-robust pose-maps to infer the 3D poses based on the results of ~\cite{openpose}.
\cite{VoluHeatmap} use volumetric heatmaps to model joint locations with an encoder-decoder network for feature compression, and a distance-based heuristic is applied to retrieve 3D pose for each person. 
\cite{zanfir2018deep} propose to group human joints according to body part scores based on integrated 2D and 3D information.
\cite{SMAP} develop 3D part affinity field for depth-aware part association, and utilizes a  refine network to refine the 3D pose given predicted 2D and 3D joint coordinates. Though avoiding repeated single person pose estimation, these methods requires a second association stage for joint grouping.

\begin{figure*}[t]
	\centering
	\vspace{-2mm}
	\includegraphics[width=0.9\textwidth]{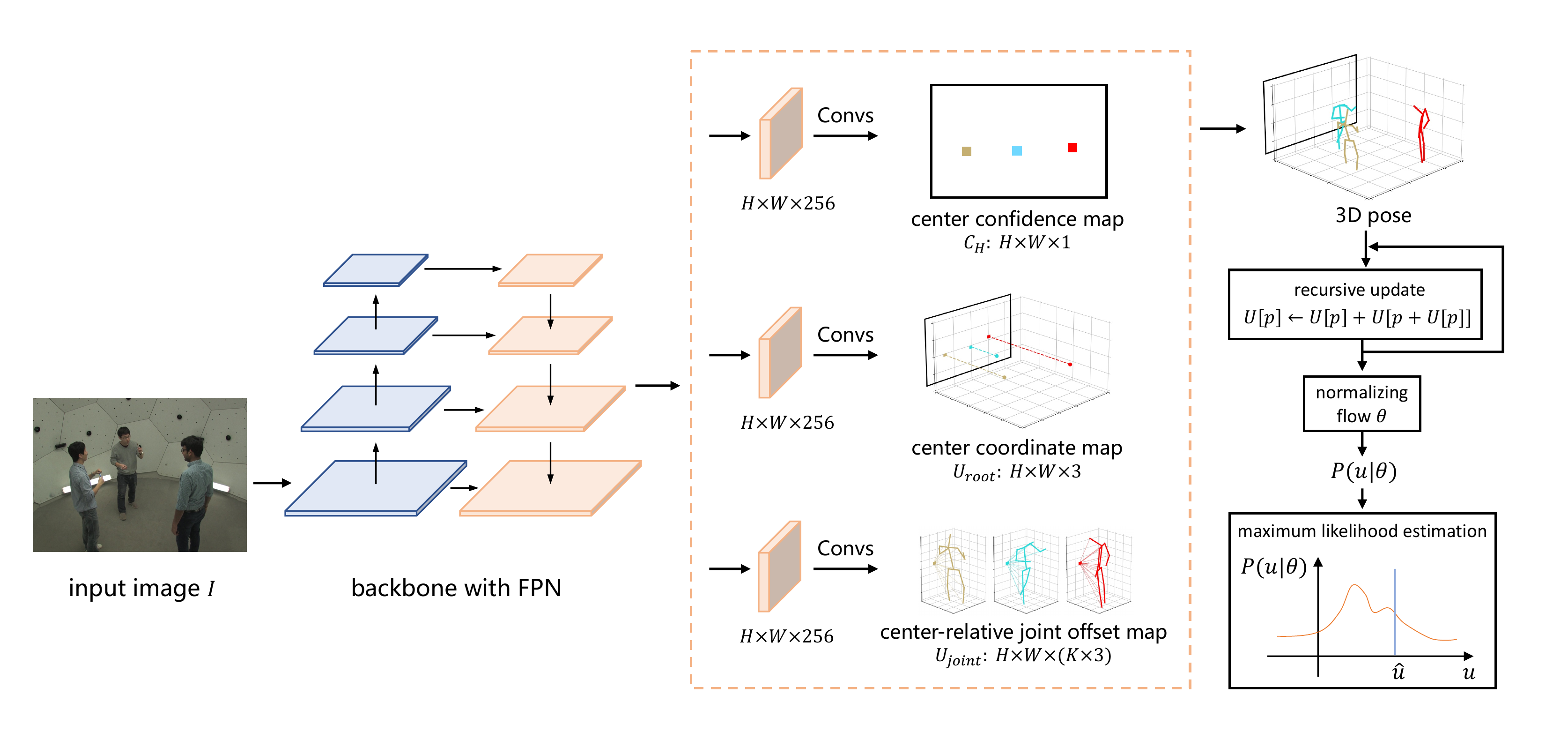}
	\vspace{-5mm}
	\caption{Illustration of the framework of DAS.
	The input image is fed to 2D CNN backbone for feature extraction. A followed FPN neck produces multi-level feature maps of different sizes. Shared prediction heads upon the multi-level FPN features are responsible for human center detection, center coordinate regression and center-relative joint offset regression respectively. 
	The 3D joint locations are reconstructed based on obtained center coordinates and joints offsets. The probability  distribution of 3D joint locations is modeled by normalizing flow with recursive update. Maximum likelihood estimation is applied to  assist the learning process.
	}
	\vspace{-2mm}
	\label{fig:framework}
\end{figure*}

\subsection{Single-stage 2D Human Pose Estimation}
There exists some researches that employ a single-stage pipeline for 2D multi-person human pose estimation.
The single-stage methods treat the pose estimation as parallel human center localizing and center-to-joint regression problem~\cite{centernet,spm,pointsetanchor}. Instead of separating joints localizing and grouping, these methods predict each of the joint offsets from the detected center points, which is usually set as the root joint of human. Since the joint offsets are directly correlated to estimated center points, this formulation avoid the manually designed grouping post-processing and is end-to-end trainable. 
Zhou et al.~\cite{centernet} direct regress the joint locations from human center. 
Nie et al.~\cite{spm} introduce a hierarchical structured pose representation to assist the long-range regression.
Wei et al.~\cite{pointsetanchor} propose to regress the joint locations from point-set anchors which serve as prior of basic human poses.
These single-stage methods have high efficiency for deployment.

\subsection{Normalizing Flows for Human Pose Estimation}
Some recent works introduce normalizing flow for 3D human pose and shape estimation.
Xu et al.~\cite{xu2020ghum} propose a statistical 3D articulated human shape model with normalizing flow representations for skeleton kinematics.
Zanfir et al.~\cite{zanfir2020weakly} introduce the kinematic latent normalizing flow representation for weakly supervised 3D human pose and shape reconstruction.
Biggs et al.~\cite{biggs20203d} and Wehrbein et al.~\cite{wehrbein2021probabilistic} propose to model the ambiguities and occlusions in 3D human pose and shape estimation by normalizing flow.
Kolotouros et al.~\cite{kolotouros2021probabilistic} learn a mapping from 2D image to a distribution of 3D body meshes based on conditional normalizing flow.
Different from the above methods, Li et al.~\cite{RLE} use normalizing flow to model the underlying distribution of joint location, and propose to optimize the parameters by residual log-likelihood estimation.

Inspired by~\cite{RLE}, we propose a recursive flow based optimization scheme for 3D joint offset regression. The distribution of joint location is modeled with normalizing flow and updated in a recursive manner. Then the regression model is optimized by maximum likelihood estimation.

\section{Method}

\subsection{Overview}
In this section, we will give an overview on the proposed Distribution-Aware Single-stage (DAS) model for multi-person 3D pose estimation. 

Given an RGB image $I$ as input, DAS estimates the joint locations  {\small $H^{img}\!=\!\{j_k^{img}\!=\!(x_k^{img}, y_k^{img}, d_k)|k\!\in\![1...K]\}$} in image coordinate system for every person. These locations are composed of the 2D image coordinates and camera-centric depth values.
DAS factorizes the multi-person 3D pose estimation as simultaneously human center localization and center-relative joint offset regression.
The illustration of the framework is shown in Figure~\ref{fig:framework}. For an input image, the CNN backbone with FPN~\cite{FPN} first extracts multi-level feature maps $\{F_1,...,F_L\}$  to deal with humans of different scales. Three parallel branches on the top of the feature maps serve for human center detection, center coordinate regression and center-relative joint offset regression respectively. The multi-person 3D poses can be reconstructed through combining the outputs of these three branches. To improve the joint localization ability in this regression based framework, a recursive flow based optimization scheme is proposed for accurately modeling the probability distribution of joint locations.
Details are given in sections below.

\subsection{Human Center Localization}

\paragraph{Ground truth assignment} Given one person with joint locations $H^{img}$, it is assigned to FPN feature map $F_l$ with downsample stride $s_l$ according to the max joint distance from root {\small $r_{max}\!=\!\max \limits_k(\sqrt{{(x_k^{img}\!-\!x_{root}^{img})^2+(y_k^{img}\!-\!y_{root}^{img})^2}})$}. Specifically, we set the regression range for $F_l$ as  $[m_{l-1}, m_l]$. If $r_{max}$ is within the range of $[m_{l-1}, m_l]$, $H^{img}$ will be assigned to the $l$th feature map. 
Then the coordinates would be scaled to {\small $H\!=\!\{j_k\!=\!(x_k, y_k, d_k)|k\!\in\![1...K]\}$} to match the downsample stride, s.t., $x_k=x_k^{img}/s_l$ and $y_k=y_k^{img}/s_l$. 
\vspace{-2mm}

\paragraph{Human center detection} We consider the root joint (i.e., pelvis) $j_{root}$ as the human center.
The center confidence map $C_H$ with shape {\small$[H\times W\times 1]$} measures whether the pixels represent any human center.
The human center detection is treated as binary classification. 
The nearest $N_{pos}$ pixels around each human center are set as positive samples $\{p_0, ..., p_{N_{pos}}\}$. The ground truth confidence is 1 for positive samples and 0 for the others. The loss function is:
\begin{equation}
    L_{cls} = \text{FocalLoss}(C_H, \hat{C}_H),
\end{equation}
where $C_H$ is the predicted center confidence map, $\hat{C}_H$ is the ground truth and FocalLoss is proposed in~\cite{retinanet}.
 Besides, a centerness branch is added to assess the center quality. 
Design of the centerness target and the loss function $L_{centerness}$ follows~\cite{FCOS}.
\vspace{-2mm}

\paragraph{Center coordinate regression}
Given the positive sample $p=(x_p,y_p)$ w.r.t the center of human $H$, the center coordinates are regressed from $p$. With the predicted center coordinate map $U_{root}$ shaped of {\small$[H\times W\times 3]$}, the regression target for the center $j_{root}=(x_{root},y_{root},d_{root})$ at $p$ is set as the offset
$U_{root}[p]=(x_{root}\!-\!x_p, y_{root}\!-\!y_p, d_{root})$.
L1 loss is adopted for this regression task:
\begin{equation}
L_{root} = \sum_p||U_{root}[p] - \hat{U}_{root}[p]||_1,
\end{equation}
where $U_{root}[p]$ is the predicted center coordinate offset at $p$ and $\hat{U}_{root}[p]$ is the ground truth.

\subsection{3D Body Joint Localization} 
\label{sec:joint_loc}

 \paragraph{Center-relative joint offset regression} 
Unlike heatmap based formulation that needs a second association stage for joint identities, this regression based formulation makes joint localization and identification a holistic process. The center-relative joint offsets are regressed from each positive sample $p$.
We predict center-relative joint offset map $U_{joint}=\{U_1,...,U_K\}$, where $U_k$ shaped of {\small$[H\times W\times 3]$} is responsible for the 3D center-relative offset of the $k$th joints. The regression target for {\small $H\!=\!\{j_k|k\!\in\![1...K]\}$} at $p$ is set as $U_k[p]=j_k-j_{root}$.
 Traditional regression loss (e.g., L1 loss) can be adopted for optimization. The objective can be written as:
\begin{equation}
L_{pose}=\sum_k\sum_p||U_k[p] - \hat{U}_k[p]||_1,
\end{equation}
where $U_k[p]$ is the predicted center-relative joint offset at $p$ and $\hat{U}_k[p]$ is the ground truth.
\vspace{-2mm}

\paragraph{Joint location distribution modeling} 
Compared to the heatmap based formulation which commonly represents the joint location as Gaussian distribution, the regression based formulation represents the joint location in a deterministic form. Denote the center-relative joint location as $u$, the deterministic representation could be more inclined to be affected by label noises, occlusions and invisibility.
Given this consideration, we employ normalizing flow~\cite{agnelli2010clustering,rezende2015variational,NICE,realnvp} to represent the center-relative joint location in a probability  distribution form: $u\sim P(u)$. 

Following~\cite{RLE}, we model the distribution $P(u)$ with reparameterization. Specifically, $P(u)$ can be obtained by scaling and shifting $z$ from a zero-mean distribution $z\sim P_Z(z)$ with transformation function $u = \bar{u} + \sigma \cdot z$, where $\bar{u}$ stands for the expectation of joint location and $\sigma$ indicates the scale of the distribution.
Given this transformation function, the density function of $P(u)$ can be calculated as:
\begin{equation}
    \text{log}~P(u) = \text{log}~P_Z(z) - \text{log}~\sigma.
    \label{eq:distr}
\end{equation}

Therefore, rather than regressing the deterministic center-relative joint location $u$, we regress the expectation $\bar{u}$ and the scale indicator $\sigma$ instead. While $P_Z(z)$ can be modeled by a normalizing flow model (e.g., real NVP~\cite{realnvp}). Note that only the expectation $\bar{u}$ is needed to calculate the eventual joint location in the inference phase.
\vspace{-2mm}

\paragraph{Recursive flow based optimization} 
Since the feature used to infer the human pose is picked from the human center position, it is less representative for joints that are far away from the center. This spatial nonalignment between feature and target can cause larger errors than the counterparts in bounding box regression~\cite{fasterrcnn,retinanet,FCOS} due to the complexity of human body structure.

\begin{figure}[t]
\centering
\begin{subfigure}{.22\textwidth}
  \centering
  \includegraphics[width=1.\linewidth]{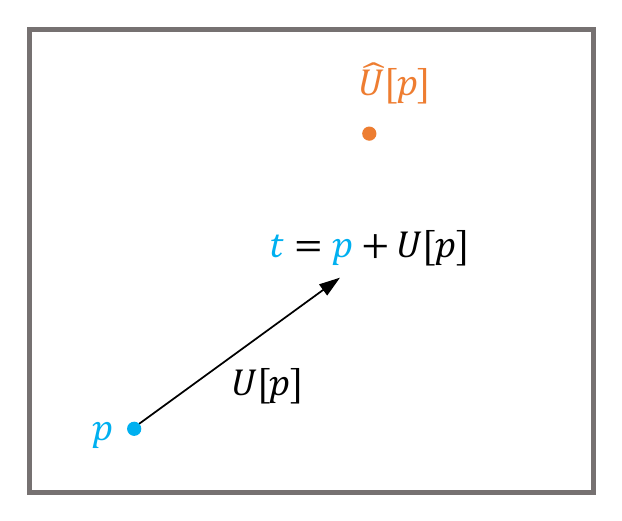}  
  \caption{initial prediction}
  \label{fig:ref1}
\end{subfigure}
\begin{subfigure}{.22\textwidth}
  \centering
  \includegraphics[width=1.\linewidth]{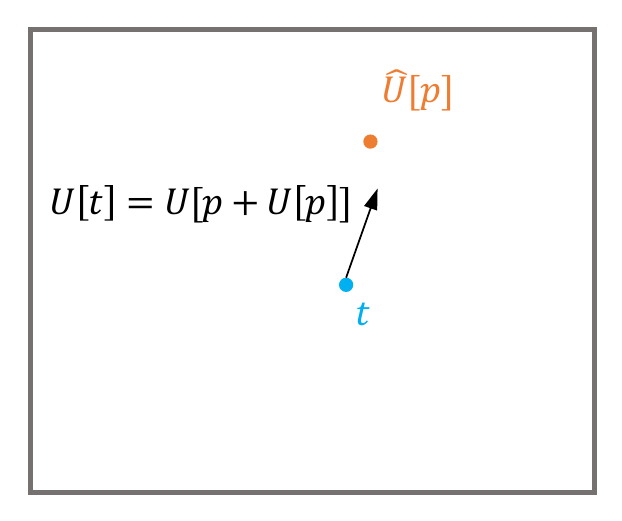}  
  \caption{successive prediction}
  \label{fig:ref2}
\end{subfigure}
\newline
\begin{subfigure}{.22\textwidth}
  \centering
  \includegraphics[width=1.\linewidth]{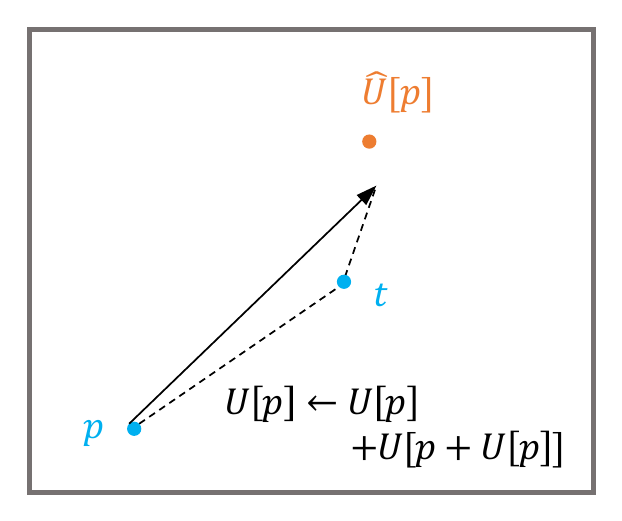}  
  \caption{recursive update}
  \label{fig:ref3}
\end{subfigure}
\begin{subfigure}{.22\textwidth}
  \centering
  \includegraphics[width=1.\linewidth]{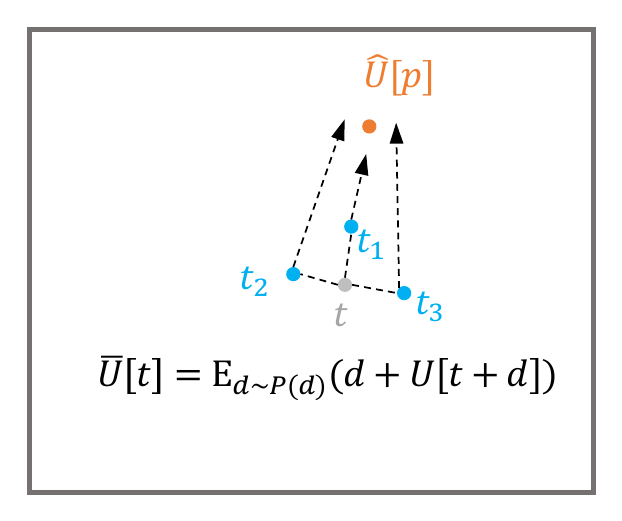}  
  \caption{multi-source update}
  \label{fig:ref4}
\end{subfigure}
\vspace{-1mm}
\caption{Illustration of the recursive update adopted in the proposed DAS model. Subscripts are omitted for simplicity.}
\vspace{-3mm}
\label{fig:update}
\end{figure}

To alleviate this problem, we propose a recursive update strategy to iteratively optimize the location expectation $\bar{u}$. The idea is depicted in Figure~\ref{fig:update}. Take the $k$th joint offset map $U_k$ as example, given the initial prediction $\bar{u}=U_k^{n}[p]$ from positive sample $p$, it is updated by the \textit{local prediction} from $p+U_p^k[p]$:
\begin{equation}
U_k^{n+1}[p] \gets U_k^n[p] + \text{B}(U_k^{local}[p + U_k^{n}[p]]),
\end{equation}
where $\text{B}(\cdot)$ indicates bilinear interpolation function to obtain values from non-integer coordinates.

For \textit{local prediction}, we can use the same prediction map as $U_k^{n}$, i.e., $U_k^{local} = U_k^n$. This formulation allows pixels around target joint to produce high quality joint offsets. By this means, $\bar{u}$ is updated in a recursive form as in Figure~\ref{fig:ref3}: 
\begin{align}
\notag
 U_k^{n+1}[p] &\gets U_k^n[p] + U_k^n[p + U_k^{n}[p]] \\
\bar{u} &\gets U_k^{n+1}[p],
\end{align}
where the interpolation operation is omitted for simplicity.

Moreover, for better modeling $\bar{u}$, we consider another multi-source update strategy to approximate the expectation. In this case, $U_k^{local}[t]$ is calculated by ensembling multiple predictions sampled around $t$ as in Figure~\ref{fig:ref4}:
\begin{align}
    \notag
  U_k^{local}[t] &= \text{E}_{d\sim P_D(d)}(d+U_k[t+d]) \\
  &\approx \sum_m P_D(d_m) (d_m + U_k[t+d_m]]),
\vspace{-2mm}
\end{align}
where $d_m$ and $P_D(d_m)$ are the sample location and probability for the $m$th sample source generated by MLP. 

The recursive update strategy can be implemented with convolutional layers and interpolation layers.
By stacking update layers on the top of joint offset map, $\bar{u}$ can be optimized progressively without modification on model pipeline. 
Besides, this recursive form avoids manually target assignment at each position in the joint offset map.

To take the advantage of the distribution-aware representation, we exploit maximum likelihood estimation for parameter optimization in the training phase.
After obtaining $\bar{u}$ and $\sigma$, joint location distribution can be expressed as in Equation~\eqref{eq:distr}.
If $P_Z(z)$ is modeled by a normalizing flow model $\theta$, the maximum likelihood estimation objective can be written as:
\begin{align}
  \notag
  L_{mle} &= -\text{log}~P(u)|_{u=\hat{u}}  \\
  &=-\text{log}~P_Z(\hat{z}|\theta) + \text{log}~\sigma,
\end{align}
where $\hat{z} = (\hat{u}-\bar{u})/\sigma$, $\bar{u}$ is the estimated expectation of center-relative joint location, and $\hat{u}$ is the ground truth location. Through $L_{mle}$, the distribution $P_Z(z|\theta)$ can be learned together with $\bar{u}$.

In this work, we further follow~\cite{RLE} to use residual log-likelihood estimation (RLE). RLE factorizes the distribution $P_Z(z)$ into one prior distribution $Q_Z(z)$ (e.g., Laplace distribution and Gaussian distribution) and one learned distribution $G_Z(z|\theta)$. The RLE objective can be written as:
\begin{equation}
    L_{rle} = -\text{log}~Q_Z(\hat{z}) -\text{log}~G_Z(\hat{z}|\theta) + \text{log}~\sigma.
\end{equation}

We recommend the reader to refer to the original paper for more details.
In experiments, we implement RLE objective to replace L1 loss for $L_{pose}$. 
\vspace{-2mm}

\paragraph{Reconstruction of 3D pose} The joint coordinates can be obtained easily by adding up the coordinates of human center and the center-relative joint offsets without other association step. Thus the 3D locations of human joints can be inferred in a single forward pass.

\begin{table*}
\centering
	\vspace{0mm}
	\begin{subtable}[t]{0.45\textwidth}
	    \renewcommand\arraystretch{1.0}
		\centering
		\resizebox{1\textwidth}{!}{
			\begin{tabu}{cccc|c}
				\tabucline[1.5pt]{-}
				center type& recursive update & MLE  & larger backbone &
				MPJPE \\
				\tabucline[1.5pt]{-}
				bbox&&&& 65.3  \\ 
				
				root&&&& 62.5  \\ 
				
				root  & \ding{51} & && 57.6 \\ 
				
				root   &&\ding{51}&&58.8\\ 
				
				root  &\ding{51} &\ding{51} && 56.3\\ 
				
				root  &\ding{51} &\ding{51} &\ding{51} & \textbf{54.4} \\
				\tabucline[1.5pt]{-}
			\end{tabu}
		}
        \caption{Component analysis of our single-stage method.}
        \label{tb:component}
	\end{subtable}
	\quad 
	\begin{subtable}[t]{0.45\textwidth}
	    \renewcommand\arraystretch{1.1}
		\centering
		\resizebox{1\textwidth}{!}{
			\begin{tabu}{ccc|c}
				\tabucline[1.5pt]{-}
				recursive update & multi-source update & update layers & MPJPE \\
				\tabucline[1.5pt]{-}
				&&& 62.5  \\ 
				
				\ding{51} && 1& 58.6  \\ 
				
				\ding{51} &\ding{51}& 1 & 58.2  \\ 
				
				\ding{51} &\ding{51}& 2 & 57.9 \\
				
				\ding{51} &\ding{51}& 3 & \textbf{57.6} \\ 
				\tabucline[1.5pt]{-}          
			\end{tabu}
		}
        \caption{Comparison of different settings for recursive update. }
    	\label{tb:update}
	\end{subtable}
	\vspace{-1mm}
	\caption{Ablation study on CMU Panoptic dataset.}
	\label{tb:abl}
\end{table*}

\begin{table*}[t]
	\renewcommand\arraystretch{1.0}
	\vspace{0mm}
	\begin{center}
		\scriptsize
		\resizebox{0.64\textwidth}{!}{
			\begin{tabu}{c||cccc|c}
			    \toprule[1pt]
				method& Haggling & Mafia & Ultimatum & Pizza & \textit{Mean} $\downarrow$\\
			    \midrule[1pt]
			    \multicolumn{6}{c}{\textit{Top-down methods}} \\
				Popa et al.~\cite{popa2017deep}&217.9&	187.3&	193.6&	221.3&	203.4 \\
				Zanfir et al.~\cite{zanfir2018monocular} &140.0	&165.9	&150.7	&156.0	&153.4 \\
				Wang et al.~\cite{HMOR} &50.9	&\textbf{50.5}	&50.7	&68.2	& 55.1*  \\
                \hline
			    \multicolumn{6}{c}{\textit{Bottom-up methods}} \\
                Zanfir et al.~\cite{zanfir2018deep}& 72.4	&78.8	&66.8	&94.3	&78.1* \\
                Fabbri et al.~\cite{VoluHeatmap}&\textbf{45}	&95	&58	&79	&69 \\
                Zhen et al.~\cite{SMAP}&63.1	&60.3	&56.6	&67.1	&61.8\\
                \hline
                \hline
                Ours& 53.3	&51.2	&\textbf{49.1} &\textbf{61.5}	&\textbf{53.8}\\
				\tabucline[1.pt]{-}    
			\end{tabu}
		}
	\end{center}
	\vspace{-3mm}
    \caption{Comparison with SOTAs on CMU Panoptic dataset in MPJPE. The top rows are two-stage methods and the bottom row is our single-stage methods. * means that \textit{Mean} MPJPE is recalculated by averaging over activities following the standard practice in~\cite{zanfir2018monocular}.}
	\label{tb:panoptic}
	\vspace{-2mm}
\end{table*}

\subsection{Training and Inference}
\paragraph{Training} 
During training, we first transform the 3D joint coordinates into image coordinate system using the camera intrinsic parameters. 
Since the same object can have different depth values in images captured from different cameras, it is hard for model to learn an absolute representation of object depth directly. Therefore we follow the previous work~\cite{3Dposenet} to use normalized depth $d_{norm}=d/f$ as target depth for center coordinate regression, where $f$ is the camera focal length. Besides, for center-relative pose offset, we do not normalize the depth value for training stabilization. 
The overall objective is:
\begin{equation}
L = L_{cls} + \lambda_1L_{centerness} + \lambda_2L_{root} + \lambda_3L_{pose},
\end{equation}
where $\lambda_1,\lambda_2,\lambda_3$ are loss weights.

\paragraph{Inference} 
The input image is first fed into the model to produce all the intermediate results. Then the positions where the center confidences are higher than a threshold (set as 0.05) are chosen as positive samples. 
Corresponding center coordinates and center-relative joint offsets are taken from the positive samples to form the camera-centric 3D poses.  Non-maximum suppression is adopted to reduce redundant pose hypothesises.

\section{Experiments}

\subsection{Datasets}

\paragraph{CMU Panoptic} 
CMU Panoptic is a large-scale real-word indoor 3D human pose dataset containing 65 video sequences of daily activities~\cite{cmupanoptic}. We follow the evaluation protocal proposed by~\cite{zanfir2018monocular} to use videos from HD cameras of indices 16 and 30. For evaluation, 9600 frames from 4 activities (\textit{Haggling}, \textit{Mafia}, \textit{Ultimatum} and \textit{Pizza}) are chosen as test set. For training, we use a mixed dataset composed of COCO~\cite{coco} and video sequences from 3 activities (\textit{Haggling}, \textit{Mafia} and  \textit{Ultimatum}). The training set has no overlap with the test set. For images with only 2D pose annotations, the depth information is ignored in loss calculation.
We follow the previous works to use  Mean Per Joint Position Error (MPJPE) for performance evaluation~\cite{zanfir2018monocular,wang2021mvp}. MPJPE is calculated after aligning the pose by the root joint.
\vspace{-2mm}

\paragraph{MuCo-3DHP and MuPoTS-3D} 
MuCo-3DHP and MuPoTS-3D are multi-person 3D pose estimation datasets for training and evalution respectively proposed in~\cite{ORPM}.
The training set MuCo-3DHP is a large-scale synthesized dataset. The images are generated by randomly compositing the persons from single person 3D pose estimation dataset MPI-INF-3DHP~\cite{3dhp}.
The test set MuPoTS-3D is a realistic dataset captured from real-world outdoor scenes which is annotated with marker-less motion capture system. MuPoTS-3D contains 20 video sequences. Each video has up to 3 subjects. 
We use the mixed datasets composed of COCO and MuCo-3DHP for training, and use MuPoTS-3D for evaluation. 
We follow the previous works to use the 3D Percentage of Correct Keypoints (3DPCK) for performance evaluation~\cite{3Dposenet,zhang2021bmp}. In particular, PCK$_{rel}$ is used to evaluate the predictions after root alignment, and PCK$_{abs}$ is used to evaluate the predictions under camera coordinate system. One joint is treated as correct if it is within 15cm distance from the matched ground truth.

\subsection{Ablation Study}
\label{sec:abl}
We conduct ablation study on CMU Panoptic dataset. The model is implemented with ResNet-50~\cite{he2016deep} and FPN~\cite{FPN}. The experimental results are shown in Table~\ref{tb:abl}.

\begin{table}
	\renewcommand\arraystretch{1.0}
	\begin{center}
		\scriptsize
		\resizebox{0.42\textwidth}{!}{
			\begin{tabu}{c||cc}
			    \toprule[1pt]
				method &$\text{PCK}_{rel} \uparrow$ & $\text{PCK}_{abs} \uparrow$ \\
			    \midrule[1pt]
			    \multicolumn{3}{c}{\textit{Top-down methods}} \\
				Rogez et al.~\cite{lcrnet++} & 70.6& n/a  \\
				Moon et al.~\cite{3Dposenet} & 81.8& 31.5 \\
				Wang et al.~\cite{HMOR} &82.0 & \textbf{43.8} \\
				Lin et al.~\cite{HDNet} &\textbf{83.7}*& 35.2* \\
				\hline
			    \multicolumn{3}{c}{\textit{Bottom-up methods}} \\
				Mehta et al.~\cite{ORPM}  &65.0& n/a \\
				Mehta et al.~\cite{xnect} & 70.4& n/a  \\
				Zhen et al.~\cite{SMAP} &73.5& 35.4\\
				\hline
				\hline
				Ours &82.7& 39.2\\
				\tabucline[1.pt]{-}    
			\end{tabu}
		}
	\end{center}
	\vspace{-4mm}
    \caption{Comparison with SOTAs on MuPoTS-3D dataset. PCK$_{rel}$ and PCK$_{abs}$ are reported for \textit{all groundtruths}. * means the value is reported on \textit{matched groundtruths}.
    }
    \vspace{-3mm}
	\label{tb:mupots}
\end{table}

We first analyse different components in our method, as shown in Table~\ref{tb:component}. Compared to the bounding box center, using root joint as human center is more beneficial for body joint regression, which brings improvement of 2.8mm in MPJPE. When equipped with recursive update strategy, 4.9mm improvement in MPJPE is obtained. 
If the joint location distribution is modeled by normalizing flow, maximum likelihood estimation (MLE) during training effectively boosts the localization ability and improves MPJPE by 3.7mm.
When combining the recursive update with MLE, MPJPE achieves 56.3mm, which exceeds baseline by 6.2mm.
With a larger backbone~\cite{mspn}, the proposed method can achieve a lower MPJPE of 54.4mm.

The detailed comparison of different settings for recursive update is shown in Table~\ref{tb:update}. By applying recursive update, MPJPE can be improved by 3.8mm compared to the baseline. With multi-source update, MPJPE is further improved by 0.4mm. Slight performance gains are achieved if stacking more update layers.

\subsection{Quantitative Results}

\begin{figure}[t]
	\centering
	\vspace{1mm}
	\includegraphics[width=0.4\textwidth]{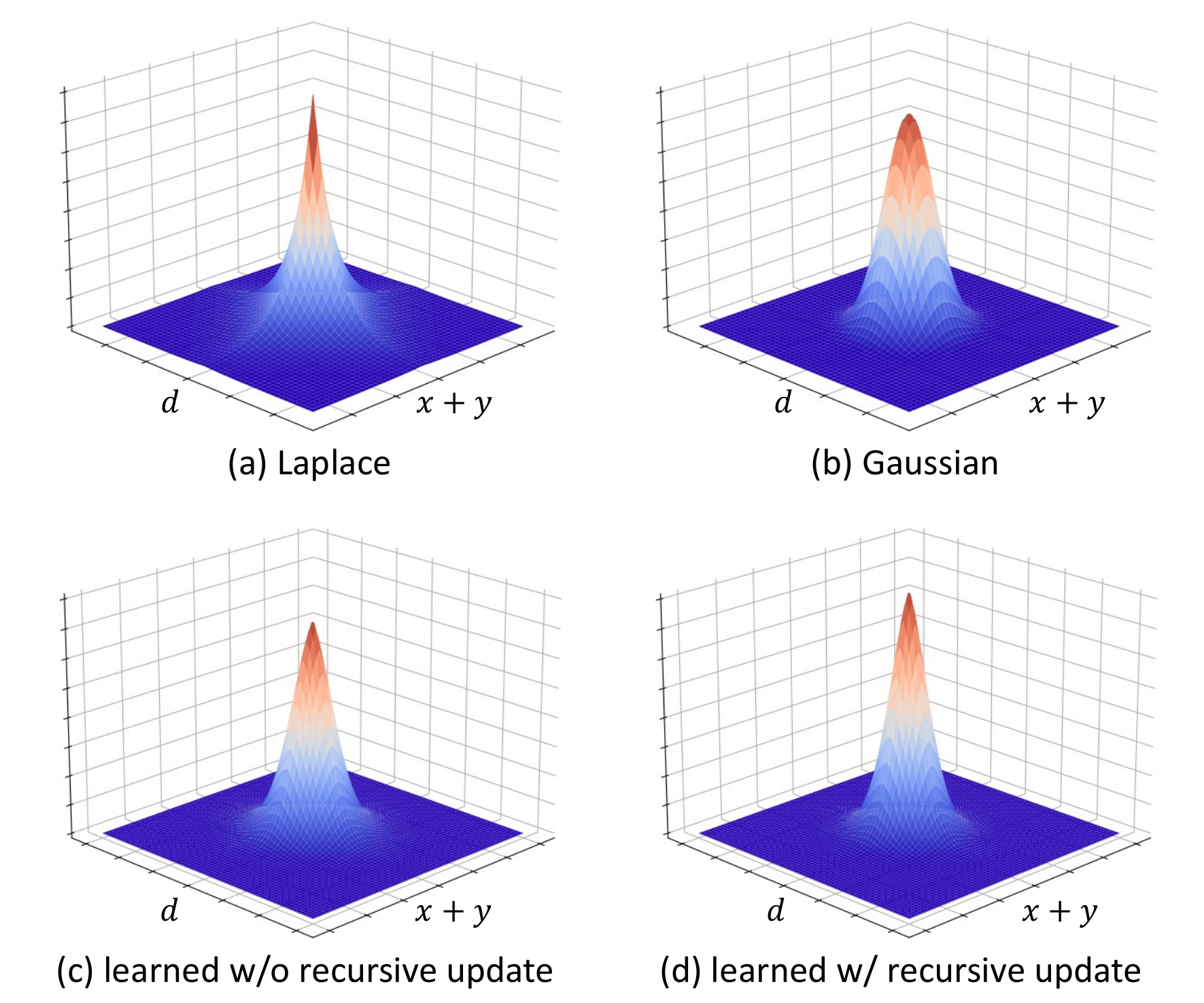}
	\vspace{-2mm}
	\caption{Visualization of the learned distribution. 
	}
	\label{fig:flow}
	\vspace{-4mm}
\end{figure}

\begin{figure*}[h]
    \vspace{-3mm}
	\centering
	\includegraphics[width=0.85\textwidth]{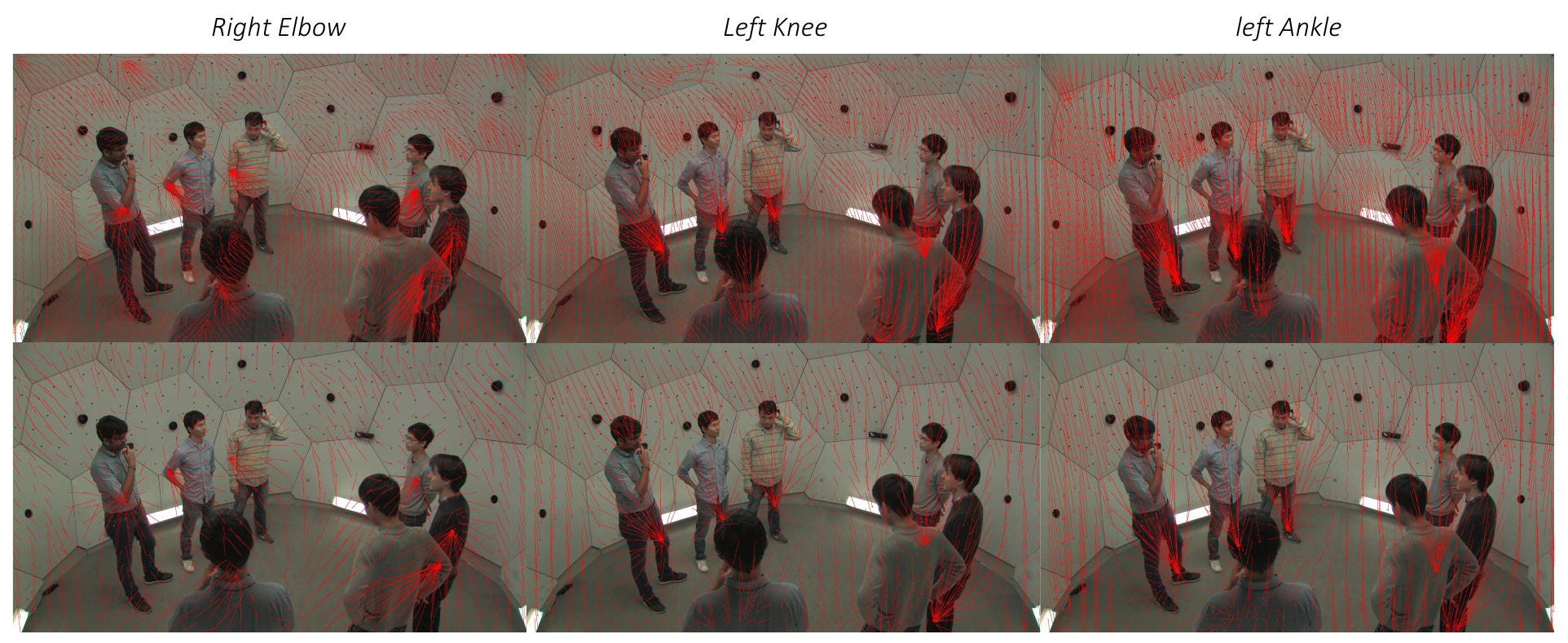}
	\vspace{-0mm}
	\caption{Visualization of the joint offsets by 2D projection. 
	The top row is taken from the prediction map with downsample stride 16 and the bottom row is taken from the prediction map with downsample stride 32. 
	}
	\vspace{-1mm}
	\label{fig:refine_vis}
\end{figure*}

\paragraph{Comparison on CMU Panpotic} 
We conduct experiments to compare with both the state-of-the-art top-down and bottom-up methods on CMU Panoptic. The quantitative results are reported in the Table~\ref{tb:panoptic}. MPJPE is adopted to evaluate the performance of 3D pose estimation after root alignment.
As shown in results, the proposed single-stage method outperforms previous volumetric based bottom-up methods in mean MPJPE, and achieves slightly better performance compared to previous top-down methods.

\paragraph{Comparison on MuPoTS-3D} 
We evaluate the proposed method on MuPoTS-3D test set using the evaluation metric PCK$_{rel}$ and PCK$_{abs}$.
We compare our method with state-of-the-art top-down and bottom-up methods in Table~\ref{tb:mupots}. 
From the experimental results, our method outperforms previous bottom-up methods especially in PCK$_{rel}$. This result demonstrates the effectiveness of our single-stage method in estimating the   human body structure. Compared to the top-down methods, our method obtains comparable  performance without a second stage for single person pose estimation and depth estimation.

\begin{table}[t]
	\renewcommand\arraystretch{1.0}
	\begin{center}
		\small
		\resizebox{0.43\textwidth}{!}{
			\begin{tabu}{c||c|c|c}
				\toprule[1.5pt]
				 method& runtime & MPJPE & PCK$_{rel}$ \\
				\midrule[1.5pt]
			    \multicolumn{4}{c}{\textit{Top-down methods}} \\
				Moon et al.~\cite{3Dposenet}* & 107~\small{ms} & n/a & 81.8 \\
				Lin et al.~\cite{HDNet}* & 118~\small{ms} & n/a& 83.7 \\
				\hline
			    \multicolumn{4}{c}{\textit{Bottom-up methods}} \\
                Fabbri et al.~\cite{VoluHeatmap} & 125~\small{ms} & 69& n/a \\
				Zhen et al.~\cite{SMAP}*&108~\small{ms}& 61.8& 73.5 \\
				\hline
				\hline
				Ours  &75~\small{ms}& 54.6 & 81.0 \\
				\tabucline[1.5pt]{-}    
			\end{tabu}
		}
	\end{center}
    \vspace{-5mm}
    \caption{Runtime comparison with SOTAs.
    * means the speed is reproduced based on official repository.}
	\label{tb:running_time}
	\vspace{-5mm}
\end{table}

\begin{figure*}[t]
	\centering
	\vspace{-0mm}
	\includegraphics[width=0.9\textwidth]{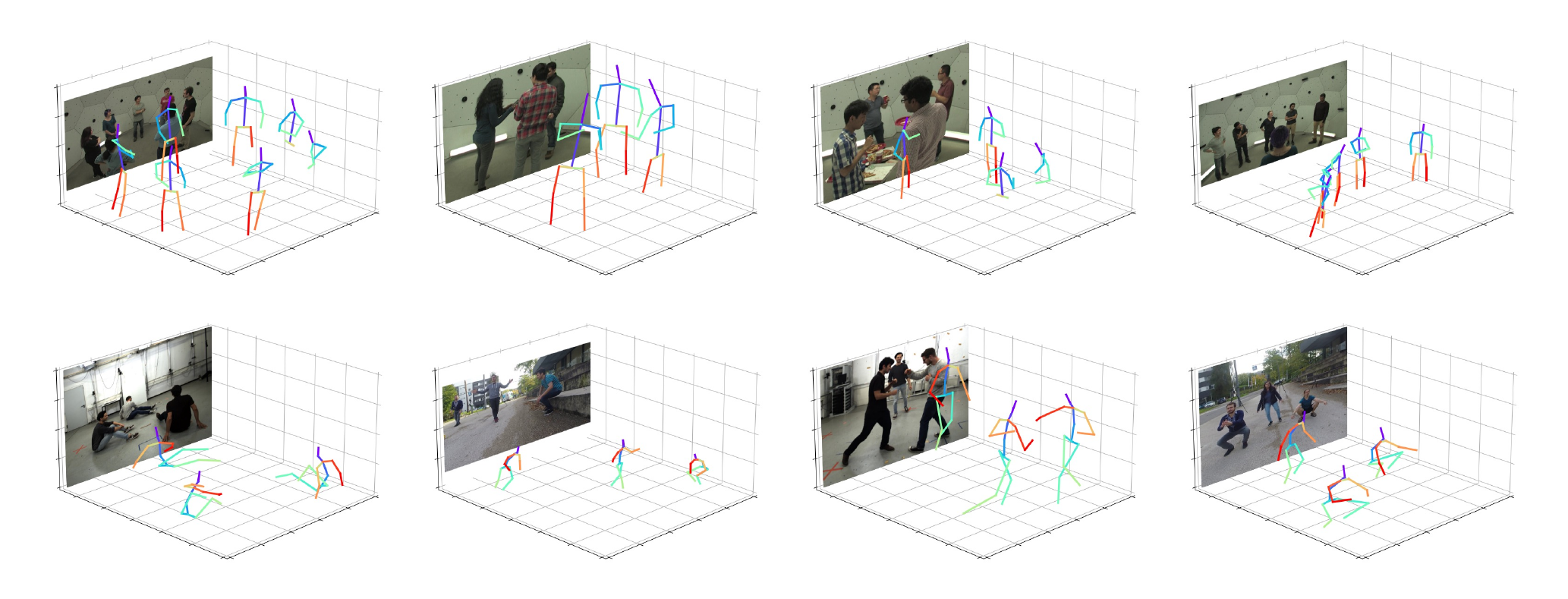}
	\vspace{-3mm}
	\caption{Qualitative results of the proposed DAS model for multi-person 3D pose estimation. Best viewed in color.
	}
	\label{fig:vis}
	
	\vspace{-5mm}
\end{figure*}

\subsection{Running Time Analysis}
We compare the running time of our method with top-down and bottom-up methods in Table~\ref{tb:running_time}. In this experiment, our method is implemented with 2-stage MSPN~\cite{mspn}, and the running time is measured on a single V100 GPU with batch size 1. Under this setting, we also give the corresponding MPJPE and PCK$_{rel}$ on CMU Panoptic and MuPoTS-3D respectively.
Compared to the top-down methods, our method achieves comparable performance with $\sim\!1/3$ less runtime. 
Considering there are few people (less than 3 per image on average) on MuPoTS-3D, the time consumption of the top-down methods can increase rapidly in complex scenes.  
Compared to the bottom-up methods, our method maintain faster inference speed since there is no need for high-resolution 2D or 3D heatmaps. This property makes DAS a computational-friendly method.

\subsection{Qualitative Results}

We visualize the distribution learned by normalizing flow in Figure~\ref{fig:flow}. The difference between learned distribution and prior distributions demonstrates the importance of distribution-aware optimization scheme. Besides, the learned distribution with recursive update has less variance than the case without, indicating the recursive update strategy assists to accurately model the  underlying distribution.
Also, we visualize the learned joint offset maps by 2D projection in Figure~\ref{fig:refine_vis}. As can be seen, with recursive flow based optimization, the pixels  learn to regress the joint location either around human center or around target joint at prediction maps. The learning of the location distribution of long-distance joints can benefit from the more precise local predictions.
In Figure~\ref{fig:vis}, we provide visualization results of our DAS model. We can see DAS can produce accurate predictions in various scenarios, e.g., pose changes, person truncation, clutter backgrounds.

\section{Conclusions}

In this paper, we present a novel  Distribution-Aware Single-stage (DAS) model for solving the challenging multi-person 3D pose estimation problem. Different from existing two-stage methods, \emph{i.e.}, top-down or bottom-up ones, the proposed DAS model can simultaneously locate the positions of persons and their body joints in camera coordinate space via a one-pass fashion. This helps to simplify the pipeline and overcome drawbacks of prior works on high computation cost and model complexity. In addition, DAS successfully introduces the normalizing flows into multi-person 3D pose estimation task and learns the joint location distribution together in the training phase. Moreover, DAS adopts a recursive flow based optimization scheme for progressively refining the location distribution. By this way, DAS derives the true underlying distribution, thus boosting the regression performance. We conduct extensive experiments on multiple benchmarks and verify the effectiveness and efficiency of the proposed DAS model for multi-person 3D pose estimation.

\vspace{-3mm}
\paragraph{Limitations}
The proposed DAS model is limited for estimating poses of extremely overlapped persons in an image. 
Since DAS represents the 3D human pose by human center and corresponding joint offsets, the overall performance is highly affected by the center detection process. When persons are extremely overlapped, their centers can also be overlapped in the image space. In this case, some occluded centers may be erroneously suppressed.  

\section*{Acknowledgments}
This research is partly supported by National Natural Science Foundation of China (Grant 62122010, 61876177), Fundamental Research Funds for the Central Universities, and Key R\&D Program of Zhejiang Province (2022C01082).

{\small
\bibliographystyle{ieee_fullname}
\bibliography{egbib}
}

\end{document}